\documentclass[10pt,twocolumn,letterpaper]{article}

\usepackage[pagenumbers]{iccv} 

\usepackage{url}
\usepackage{wrapfig}
\usepackage{graphicx}
\usepackage{multirow}
\usepackage{siunitx}
\usepackage{booktabs}
\usepackage{adjustbox}
\usepackage{tabularx}
\usepackage{colortbl}
\usepackage{xcolor}
\usepackage{float}

\definecolor{best}{rgb}{0.96, 0.57, 0.58}
\definecolor{second}{rgb}{0.98, 0.78, 0.57}
\definecolor{third}{rgb}{1.0, 1.0, 0.56}

\usepackage{algorithm}
\usepackage{algpseudocode}

\definecolor{iccvblue}{rgb}{0.21,0.49,0.74}
\usepackage[pagebackref,breaklinks,colorlinks,allcolors=iccvblue]{hyperref}
\usepackage{graphicx}
\usepackage{tikz}
\usepackage{comment}

\title{UniGeo: Taming Video Diffusion for Unified Consistent Geometry Estimation}

\author{
    Yang-Tian Sun$^{1}$ 
    \quad Xin Yu$^{1}$ 
    \quad Zehuan Huang$^{2}$
    \quad Yi-Hua Huang$^{1}$\\
    \quad Yuan-Chen Guo$^{3}$
    \quad Ziyi Yang$^{1}$
    \quad Yan-Pei Cao$^{3}$
    \quad Xiaojuan Qi$^{1\dagger}$ 
    \\
    $^1$The University of Hong Kong \quad
    $^2$Beihang University \quad
    $^3$VAST
}

\begin{document}
\twocolumn[{%
\renewcommand\twocolumn[1][]{#1}%
\maketitle

\begin{center}
    \centering
    \captionsetup{type=figure}
    \includegraphics[width=1\linewidth, trim=0 0 0 0, clip]{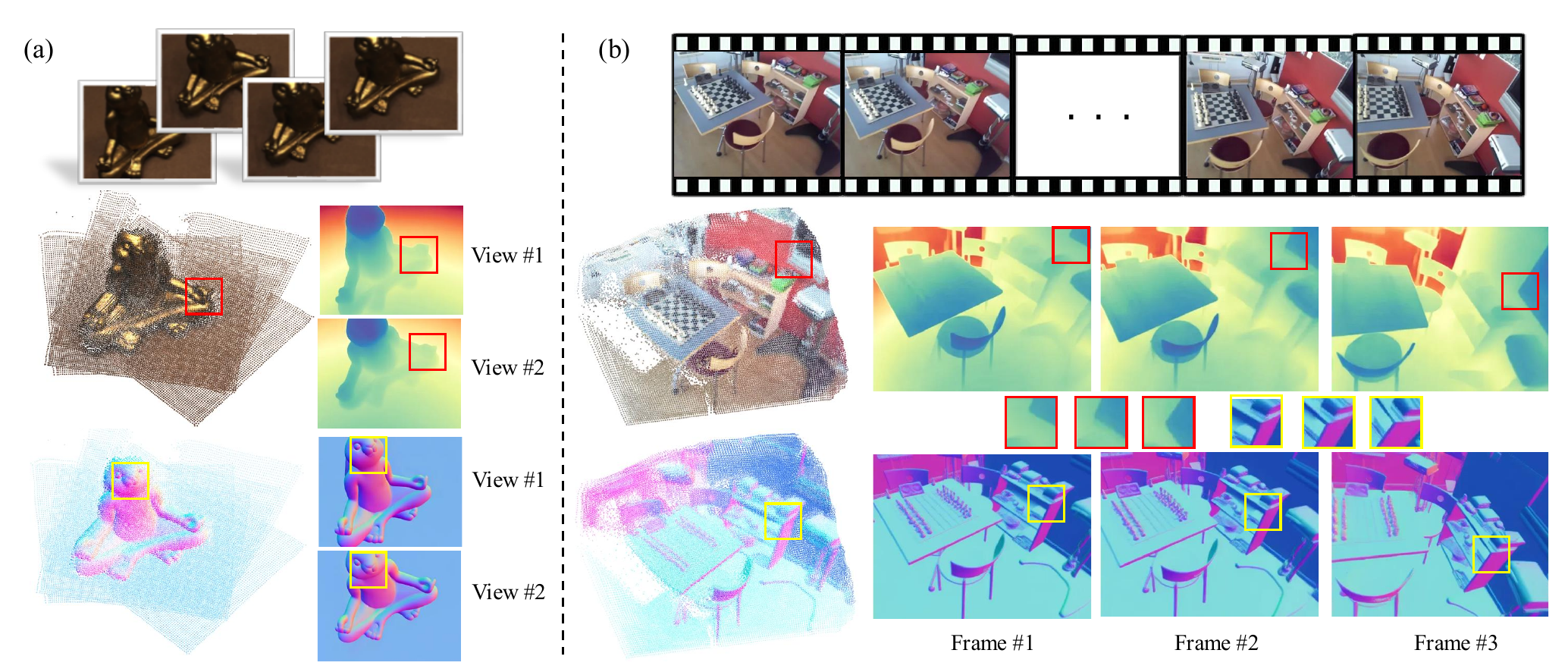}
    \vspace{-6mm}
    \captionof{figure}{
    UniGeo utilizes video diffusion models to jointly estimate geometric properties—such as surface normals and coordinates-from either multi-view images (a) or video sequences (b). Rather than predicting within the local camera coordinate system of each frame, UniGeo infers geometric attributes in a unified global reference frame. Such design facilitates consistent estimation across frames for patches corresponding to the same 3D region by effectively leveraging inter-frame patch correspondences embedded in the video prior. Moreover, the estimated properties can be seamlessly integrated into downstream tasks such as 3D reconstruction.
    Project page: \href{https://sunyangtian.github.io/UniGeo-web/}{https://sunyangtian.github.io/UniGeo-web/}
    }
    \vspace{-1mm}
\label{fig:teaser}
\end{center}
}]

\begin{abstract}
Recently, methods leveraging diffusion model priors to assist monocular geometric estimation (e.g., depth and normal) have gained significant attention due to their strong generalization ability. However, most existing works focus on estimating geometric properties within the camera coordinate system of individual video frames, neglecting the inherent ability of diffusion models to determine inter-frame correspondence.
In this work, we demonstrate that, through appropriate design and fine-tuning, the intrinsic consistency of video generation models can be effectively harnessed for consistent geometric estimation. Specifically, we 1) select geometric attributes in the global coordinate system that share the same correspondence with video frames as the prediction targets, 2) introduce a novel and efficient conditioning method by reusing positional encodings, and 3) enhance performance through joint training on multiple geometric attributes that share the same correspondence. Our results achieve superior performance in predicting global geometric attributes in videos and can be directly applied to reconstruction tasks. Even when trained solely on static video data, our approach exhibits the potential to generalize to dynamic video scenes.

\end{abstract}    
\section{Introduction}
\label{sec:intro}

Estimating 3D geometric information, such as depth and surface normals, from RGB input frames is a fundamental task in computer vision with applications spanning VR/AR, robotics, and autonomous driving. Recently, single-image 3D geometry estimation has attracted significant attention~\cite{Ke2023RepurposingDI, fu2024geowizard, piccinelli2024unidepth, yin2023metric3d, yang2025depth, yang2024depth}. Approaches like Marigold~\cite{Ke2023RepurposingDI} and Geowizard~\cite{fu2024geowizard} have demonstrated that diffusion-based image generators, when fine-tuned, can achieve remarkable performance in depth and normal prediction tasks. These findings suggest that priors learned by image generation models from large-scale datasets can enhance the accuracy and generalizability of geometric estimations.

\begin{figure}
    \centering
    \includegraphics[width=1.0\linewidth]{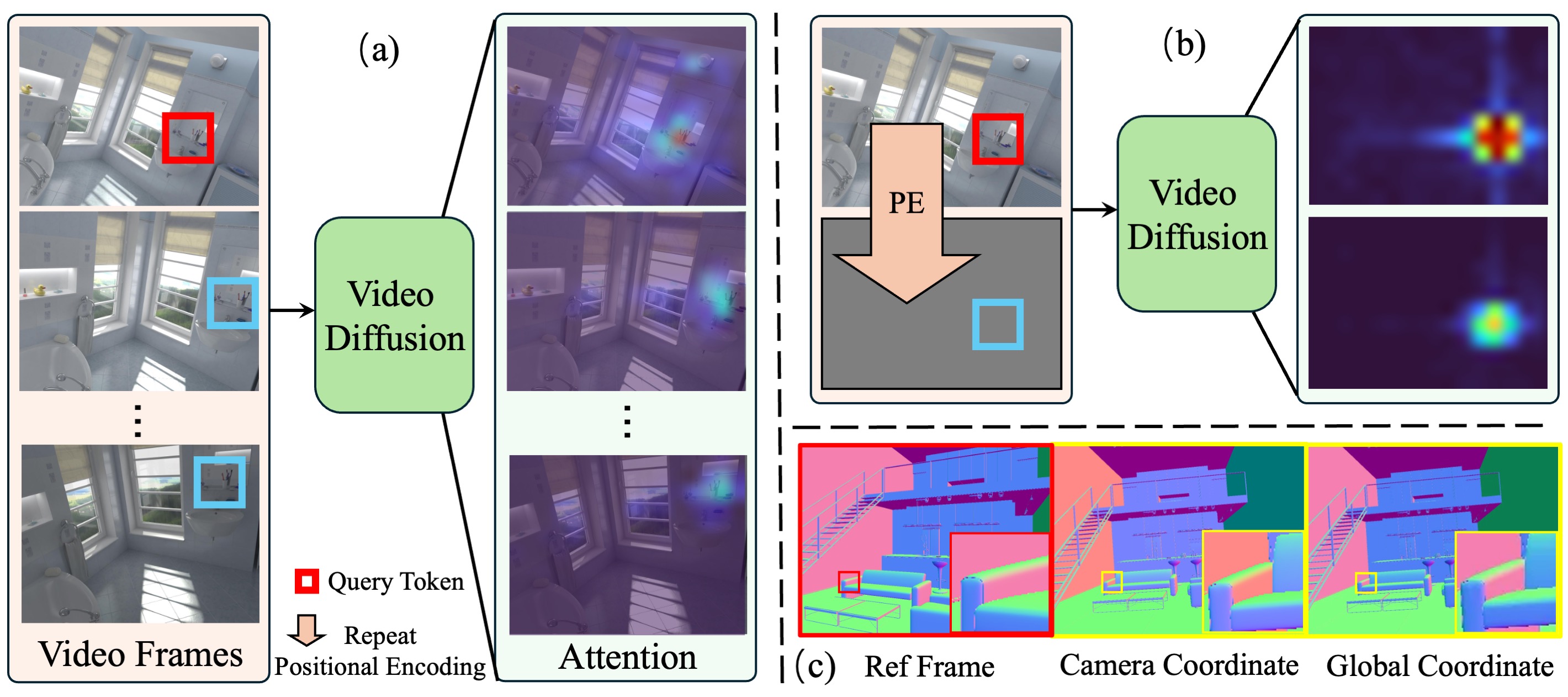}
    \vspace{-6mm}
    \caption{
    Our key insights lie in (a) pre-trained video diffusion models capture accurate inter-frame correspondence (the same patches in different frames highlight in the attention maps), (b) the correspondence can be specified by applying identical positional encodings onto different frames, and (c) geometric properties within a shared global coordinate system naturally exhibit alignment across frames.
    }
    \label{fig:motivation}
    \vspace{-6mm}
\end{figure}

However, directly applying image-based geometric estimation methods to videos in a frame-wise manner often leads to noticeable inconsistencies. To mitigate this issue, recent works~\cite{hu2024depthcrafter, Shao2024LearningTC} have explored leveraging consistency priors from video diffusion models for depth estimation, treating video frames as conditioning inputs while predicting depth across frames in camera coordinates as the output.
Despite these efforts, the consistency required for geometric properties such as depth and surface normals differs fundamentally from that of RGB video frames. For instance, video priors typically enforce appearance to be similar for the same object across frames, whereas its depth and normal vary according to camera motion (see Fig.~\ref{fig:motivation} (c)). This discrepancy can lead to inaccurate geometric predictions.
Furthermore, RGB conditioning is introduced into video diffusion models via channel concatenation, altering the input format compared to the pretrained model. This necessitates architectural modifications and is hard to fully exploit the potential of video diffusion priors.

In this paper, we introduce \textit{UniGeo}, a unified framework that reformulates video-based geometry estimation tasks-- including global position and surface normals-- as a video generation problem. 
Our key insight is grounded in the discovery that pre-trained video generation models inherently possess the capability to extract inter-frame consistency, as can be visualized through attention weights across tokens, illustrated in Fig.~\ref{fig:motivation}(a). Such correspondence motivates us to repurpose a pretrained video diffusion model for consistent video geometry estimation. 

First, to better exploit consistency priors, we propose representing geometric properties within a shared global coordinate system. This approach naturally aligns geometric correspondences across frames, mirroring the consistency in RGB videos (see Fig.~\ref{fig:motivation}(c)). In contrast, prior methods~\cite{hu2024depthcrafter, Shao2024LearningTC, video_depth_anything} estimate geometry in camera-centric coordinates, which inherently introduces inconsistencies.



Second, instead of stacking RGB inputs in the channel dimension as conditions-- an approach that misaligns with the pretrained video diffusion model-- we propose treating RGB frames as additional inputs within a unified video sequence. Specifically, we organize them alongside the noised geometry sequence, enabling direct adaptation of video diffusion models without architectural modifications (refer to Fig.~\ref{fig:pipeline}). Then, motivated by the observation that attention weights between tokens naturally capture inter-frame correspondences, with these weights strongly influenced by token positional embeddings (see Fig.~\ref{fig:motivation}), we propose a novel shared positional encoding strategy that reuses positional embeddings from images and applying them to geometric properties. This achieves precise conditioning from images to geometric properties and effectively harnesses the pretrained model’s inter-token correspondence learning for improved geometry estimation.

Finally, to effectively utilize available training datasets for learning generalized models, we explore training a single network to predict multiple geometric attributes simultaneously.
Our novel formulation enables these tasks to share the same learned correspondences, allowing them to mutually reinforce each other. Surprisingly, experimental results demonstrate that this multi-task approach not only offers the added advantage of inferring multiple attributes within a unified model but also \textit{outperforms individually trained networks for specific tasks}.

\begin{figure}
    \centering
    \includegraphics[width=0.95\linewidth]{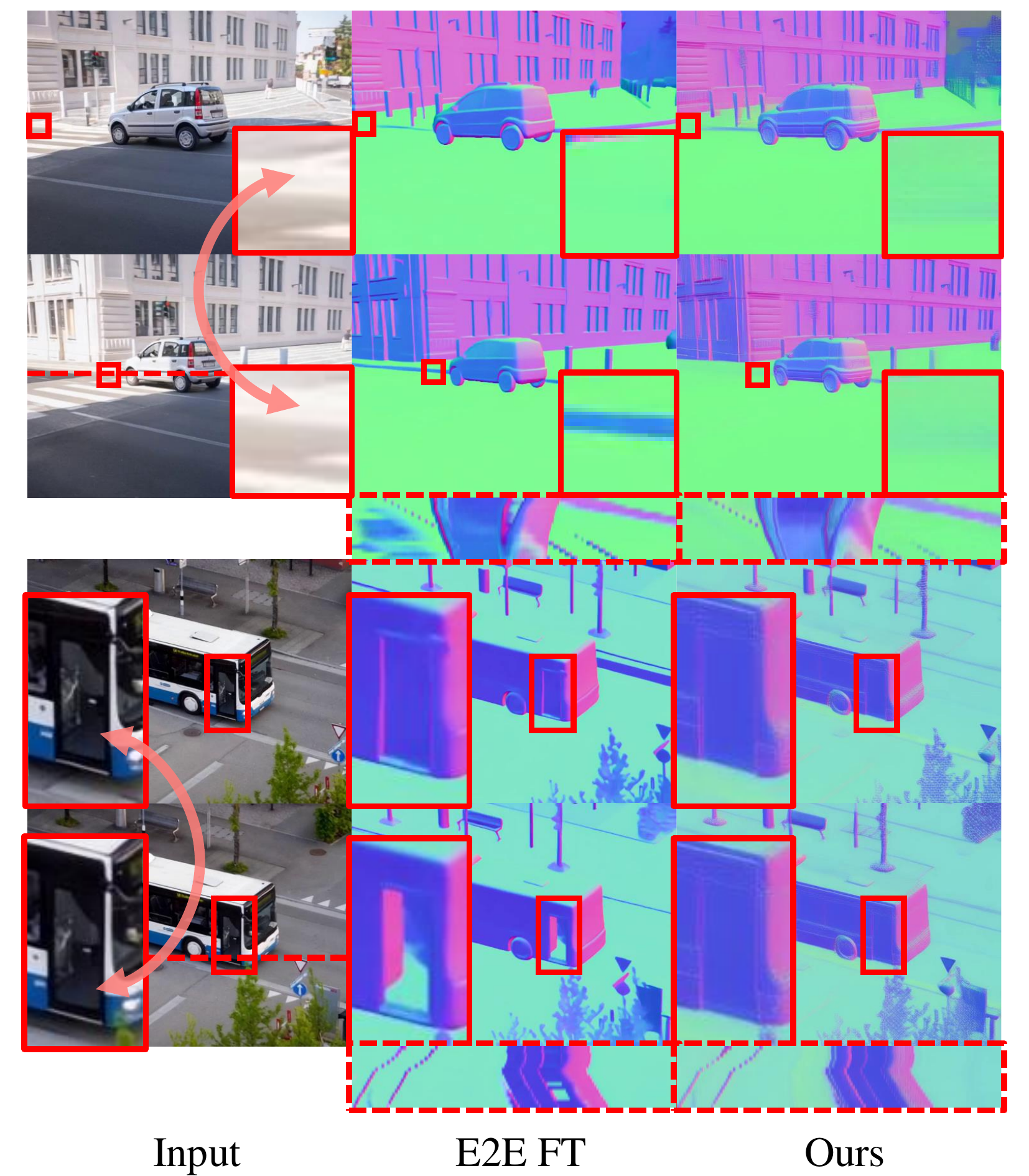}
    \caption{\textbf{Comparison to E2E.} Our method predicts normals of better consistency with more details preserved .}
    \label{fig:video_normal_dynamic}
    \vspace{-3mm}
\end{figure}

To the best of our knowledge, our work is the first capable of simultaneously predicting multiple geometric attributes (e.g., radius, normals) from video data, ensuring global consistency suitable for direct reconstruction tasks (see Fig.~\ref{fig:teaser}). Compared to image-based methods, our approach achieves superior performance without additional camera information (Table~\ref{tab:video_normal_depth_comparition}), and delivers reconstruction quality comparable to models trained on large-scale datasets (Table~\ref{tab:video_recon}). Notably, despite being trained exclusively on static data, our model benefits from video diffusion priors, enabling robust generalization to certain dynamic scenes (Fig.~\ref{fig:video_normal_dynamic}).
In summary, our contributions are: 
\begin{itemize}
    \item We propose a unified formulation of video-based geometry estimation as a video generation task, enabling the direct use of pretrained video diffusion models to achieve consistent predictions across frames. 
    \item  We introduce global coordinate representation and a novel RGB conditioning method with a shared positional encoding strategy, which allow pretrained video diffusion models to transfer learned consistency priors without requiring architectural modifications. 
    \item  We explore a multi-task learning approach that harnesses shared knowledge across tasks, enabling a unified model to simultaneously predict multiple geometric attributes from videos. 
    \item We demonstrate that our approach improves consistency and accuracy across geometric tasks and achieves competitive performance compared to state-of-the-art methods on geometry prediction and reconstruction.
\end{itemize}


\section{Related Work}
\label{sec:related_work}

\subsection{Monocular Geometry Estimation}

Early works~\cite{eigen2014depth,shelhamer2015scene} used CNNs to estimate depth from annotated datasets~\cite{geiger2013vision,silberman2012indoor}. Depth and normal were soon jointly predicted due to their interdependence~\cite{eigen2015predicting,li2015depth,qi2018geonet,qi2020geonet++}. Early works' reliance on small datasets hindered generalization to new domains.
Larger datasets proved critical for improving generalization~\cite{chen2016single,xian2018monocular,eftekhar2021omnidata}. MiDas~\cite{ranftl2020towards} introduced an affine-invariant loss for depth estimation, enabling effective training across diverse datasets.
MoGe~\cite{wang2024moge} further improved the loss design for coordinate prediction. 
Enhanced strategies like probabilistic modeling~\cite{bae2021estimating} and iterative refinement~\cite{bae2024rethinking} also improved normal estimation.

With larger annotated datasets, powerful architectures became crucial. Vision Transformers (ViTs) were applied to depth~\cite{ranftl2021vision, piccinelli2024unidepth} and normal estimation~\cite{bae2024rethinking}. Depth Anything (DA)~\cite{yang2024depth,yang2025depth} showed that ViTs trained with synthetic data can preserve depth details.
Diffusion models~\cite{sohl2015deep} emerged as scalable architectures for image generation~\cite{rombach2022high,ho2020denoising,song2020score} and proved effective for geometry estimation~\cite{xu2024matters}. Marigold~\cite{ke2024repurposing} fine-tuned SD's U-Net~\cite{rombach2022high} for high-quality depth estimation. DepthFM~\cite{gui2024depthfm} improved efficiency by reducing sampling steps via flow matching.
GeoWizard~\cite{fu2024geowizard} and DMP~\cite{lee2024exploiting} utilized diffusion priors for depth and normal estimation, while StableNormal~\cite{ye2024stablenormal} refined normals through iterative diffusion. GenPercept~\cite{xu2024matters} analyzed pre-trained diffusion models, offering insights for advancing monocular diffusion-based perception.

\subsection{Video / Multi-view Geometry Estimation}

DUSt3R~\cite{wang2024dust3r} introduces a dual ViT architecture to predict dense geometry from image pairs. Subsequent works extend DUSt3R to handle multi-view images or videos using techniques such as spatial memory~\cite{wang20243d}, multi-view fusion~\cite{tang2024mv}, zero-convolution~\cite{lu2024align3r}, and recurrent neural networks~\cite{wang2025continuous}. To enhance DUSt3R for dynamic scenes, MonST3R~\cite{zhang2024monst3r} separates the supervision of dynamic foreground and static background. Stereo4D~\cite{jin2024stereo4d} leverages stereo videos to annotate 3D tracked points and trains a time-dependent DUSt3R.
Several works extend DA to video geometry. Prompt Depth Anything~\cite{lin2024prompting} uses LiDAR-based low-resolution depth maps as prompts for accurate video depth estimation. Video Depth Anything~\cite{chen2025video} adds temporal layers to DA for relative video depth prediction. 

Diffusion models also demonstrate exceptional performance in video generation. Commercial products such as SORA~\cite{brooks2024video}, Pika~\cite{pika}, Keling~\cite{keling}, and Hailuo~\cite{hailuo} have revolutionized media creation. Open-source projects like Stable Video Diffusion (SVD)~\cite{blattmann2023stable} and CogVideo~\cite{hong2022cogvideo} leverage Unet~\cite{ronneberger2015u} with temporal attention to extend capabilities from image generation to video.
With the DiT~\cite{peebles2023scalable} architecture showcasing superior scalability, implementations like HunyuanVideo~\cite{kong2024hunyuanvideo} and CogVideoX~\cite{Yang2024CogVideoXTD} have achieved remarkable results.
Building on the success of video diffusion models, video geometry estimation methods have emerged by leveraging rich learned priors. DepthCrafter~\cite{hu2024depthcrafter} predicts video depth using pre-trained image-to-video priors, while ChronoDepth~\cite{Shao2024LearningTC} refines depth estimation with a fine-tuned SVD. 
Finetuning a video diffusion model to predict geometry attributes with diverse supervision modalities could scale to larger datasets, which are underexplored.

\begin{figure*}[t]
    \centering
    \includegraphics[width=1\linewidth]{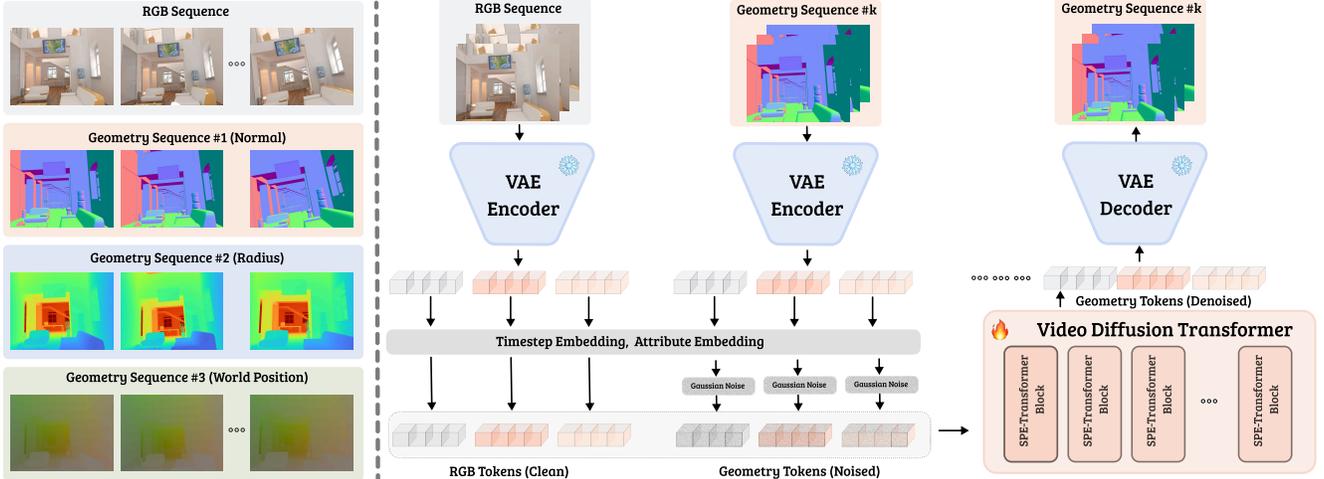}
    \vspace{-8mm}
    \caption{\textbf{Method overview.} Our method targets at predicting geometric properties that are defined in the global coordinate system, where ``radius'' represents the distance from a 3D points to the origin (left). We efficiently adapt a pre-trained video diffusion model that inherently encodes inter-frame correspondence into a consistent geometry estimation model (right), where we process both rgb sequence and geometry sequence through our proposed SPE-Transformer.}
    \label{fig:pipeline}
    \vspace{-6mm}
\end{figure*}

\section{Preliminaries}
\label{sec:preliminary}

\textbf{Diffusion Models.} Diffusion models~\cite{Ho2020DenoisingDP} can model a specific data distribution through an iterative denoising process. Specifically, Gaussian noise at different levels $t \in \{1, \ldots, T\}$ is progressively added to the data point $x_0$ in the forward process, generating a noisy sample sequence $\{x_t\}_{t=1}^T$, formulated as 
\begin{equation}
    x_t = \sqrt{\bar{\alpha}_t} x_0 + \sqrt{1 - \bar{\alpha}_t} \epsilon ,
\end{equation}
where $\epsilon \in \mathcal{N}(0, I)$, $\bar{\alpha}_t := \prod_{s=1}^t (1-\beta_s)$, and $\beta_1, \ldots, \beta_T$ is the variance schedule of a process with T steps. The denoising model $\epsilon_{\theta}(\cdot)$ parameterized with parameters $\theta$ aims to gradually reverse this process by modeling the probability $p_\theta(x_{t-1} | x_t)$. 

During the training phase, the model takes noisy data sample $x_t$ and  timestep $t$ as input, and predicts the noise $\hat{\epsilon} = \epsilon_{\theta}(x_t, t)$. Paramters 
$\theta$ is updated by minimizing the following objective function
\begin{equation}
    \mathcal{L} = \mathbb{E}_{x_0, \epsilon \sim \mathcal{N}(0,I), t \sim \mathcal{U}(T)} || \epsilon - \hat{\epsilon}||_2^2 .
\end{equation}
At the inference phase, $x_0$ is obtained by iteratively denoising a sampled Gaussian noise.

To reduce the computational cost of high-resolution inputs, Latent Diffusion Model~\cite{Rombach2021HighResolutionIS} (LDM) is often adopted by using a pre-trained VAE to encode the data into a latent space for probability modeling.

\textbf{Video Diffusion Models.} 
Given an RGB video of shape $H \times W \times F \times 3$, it is first compressed into the latent space using a pre-trained VAE encoder, obtaining a latent representation of shape $h \times w \times f \times c$. Typically, diffusion models are implemented using a U-Net architecture~\cite{Ronneberger2015UNetCN}. Recently, Diffusion Transformers (DiT)~\cite{Peebles2022ScalableDM} have demonstrated significant potential due to their superior generation quality and greater flexibility. The DiT architecture applies a patchify operation to video latent representations, converting them into tokens, which are then concatenated into a long sequence for denoising.
Our approach is based on the DiT architecture video diffusion~\cite{Yang2024CogVideoXTD}.

\section{Method}
\label{sec:method}

Our primary goal is to efficiently adapt a pre-trained video diffusion model into a video-based consistent geometry estimation model.
The key insight is that pre-trained video generation models inherently encode inter-frame correspondence, which can be leveraged for consistent video geometric property estimation through appropriate design.
In Section~\ref{sec:geometry_coord}, we propose predicting geometric properties consistent with video frames in a global coordinate system.
In Section~\ref{sec:multi-attr}, we propose a unified joint-training and inference framework for multiple geometry tasks to improve generalization. In Section~\ref{sec:shared_positional_encoding}, we introduce a conditioning method with a novel shared positional encoding strategy to enhance RGB-geometry alignment and inter-frame consistency.
In Section~\ref{sec:training_protocal}, we accelerate training and inference by employing single-step deterministic training.

\subsection{Geometry in Global Coordinate System}
\label{sec:geometry_coord}
Video generation models can produce consistent and temporally coherent video frames. Although some existing works have explored geometric estimation (e.g., depth estimation~\cite{Hu2024DepthCrafterGC, Shao2024LearningTC}) based on video diffusion, these methods focus on per-frame estimation in their respective camera coordinate systems. As a result, they fail to transfer inter-frame consistency to geometric properties.

Unlike existing per-frame prediction methods, we find that when geometric properties are defined in a global coordinate system, their geometric measurements exhibit a high correlation with color measurements—for example, the same point in space consistently appears with similar colors across different video frames. This aligns well with the consistency priors inherent in video generation models.

Therefore, in our design, we directly predict geometric properties such as position and normal in a global coordinate system across all frames, which is proven to achieve more accurate results compared to predicting in their own coordinate systems in Sec~\ref{sec:ablation}. Additionally, our approach offers the advantage that \textit{\textbf{the predicted global results can be directly used for reconstruction without requiring camera information as input}}.

\subsection{Multi-Attributes Joint Training}
\label{sec:multi-attr}

Our method focuses on leveraging the inter-frame consistency of RGB frames to learn globally consistent geometry. For different geometric attributes, such as position and normal, since they share the exact same correspondence, they can be effectively integrated and trained together. Therefore, we employ a unified training framework by utilizing multiple RGB-geometry data pairs, as shown in \cref{fig:pipeline} (left). Formally, given a collection of datasets $\mathcal{D} = \{\mathcal{D}_k\}_{k=1}^{K}$, where each dataset $\mathcal{D}_k = \{(I_j, G_j)\}_{j=1}^{M_k}$ contains two components:
 an RGB video sequence $I_j \in \mathbb{R}^{H \times W \times F \times 3}$, and  
 a corresponding geometry sequence $G_j \in \mathbb{R}^{H \times W \times F \times d}$,
where $H$ and $W$ represent spatial dimensions, $F$ denotes the number of frames, and $d$ indicates the channel dimension of the geometry attribute (e.g., $d = 3$ for normals). We train the  model on these multiple datasets with an attribute identifier $k$ (indicating, for instance, normals or positions). The attribute identifier $k$ is integrated directly into the Transformer layers to explicitly guide the model toward the desired geometric property. At inference time, the attribute label \( k \) can be specified to control the desired geometric estimation type, enabling a single model to handle depth estimation, normal prediction, and other geometric tasks within a unified framework.

\textbf{Leveraging Multi-view Data. }
Finetuning Video Diffusion requires paired video data with corresponding geometric property labels. However, such data is extremely scarce in practice. Note that the correspondence between video frames and their geometric properties also applies to multi-view images, we propose a mixed training strategy to fully utilize existing high-quality multi-view datasets, such as Hypersim~\cite{Roberts2020HypersimAP}. Specifically, we first group the data based on the bidirectional overlap between multi-view frames. For multi-view images within the same group, their latents are obtained by individually encoding each image using the video VAE, following a method similar to ~\cite{Singer2022MakeAVideoTG, Blattmann2023StableVD}. Please refer to the supplementary material for more details regarding to the dataset grouping.

\subsection{Shared Positional Encoding (SPE)}
\label{sec:shared_positional_encoding}

\textbf{Conditioning by Extending Tokens.}
A common approach for conditional generation in video models is channel-wise concatenation~\cite{Ke2023RepurposingDI, fu2024geowizard}. However, this requires modifying the original network architecture and introducing new parameters, which harms pre-trained prior and shows limited performance (see \cref{tab:ablation}). In contrast, we propose directly treating the conditional RGB sequence as extended frames, thus eliminating the need to modify the network architecture. Formally, as shown in \cref{fig:pipeline} (right), given the pre-trained VAE encoder $Enc$, we extract the RGB tokens $z^{\text{rgb}}=Enc(I_j)\in\mathbb{R}^{h \times w \times f \times c}$ and geometry tokens $z^{\text{geo}}=Enc(G_j)\in\mathbb{R}^{h \times w \times f \times c}$. After adding noise to the target geometry tokens, i.e., $z^{\text{geo}}_t = \sqrt{\bar{\alpha}_t} z^{\text{geo}}_0 + \sqrt{1 - \bar{\alpha}_t} \epsilon$ (see \cref{sec:preliminary}), we concatenate their VAE tokens along the token dimension, jointly treating them as an extended token sequence:
\begin{equation}
z^{\text{input}}_t = [z^{\text{rgb}}; z^{\text{geo}}_t],
\end{equation}
where $[\,;\,]$ denotes concatenation along the token dimension.
Subsequently, in the forward process of the DiT network, self-attention is applied across the entire sequence, enabling full feature exchange. To obtain the predicted denoised geometry result, we only retain the final half of the tokens from the output sequence and decode them to the pixel space using the VAE decoder.

\textbf{Shared Positional Encoding (SPE).} 
To fully utilize the learned inter-frame consistency in video diffusion, we further propose a \textit{Shared Positional Encoding (SPE)} strategy without changing the network architecture. Specifically, as illustrated in Fig.~\ref{fig:motivation}, we observe that attention weights among tokens strongly correlate with their positional embeddings: by repeating one frame's positional embeddings to another, different tokens corresponding to the same positional embedding consistently exhibit significantly higher mutual attention weights.
Motivated by this insight, we propose to explicitly reuse the positional embeddings from RGB tokens for geometry tokens. 

Formally, let $p_i^{\text{rgb}}$ and $p_i^{\text{geo}}$ denote the positional embeddings associated with the $i$-th RGB and geometry tokens, respectively. During training and inference, we discard the original geometry positional embeddings $p_i^{\text{geo}}$ and replace them with the RGB positional embeddings $p_i^{\text{rgb}}$:
\begin{equation}
p_i^{\text{geo}} \leftarrow p_i^{\text{rgb}}, \quad \forall i.
\end{equation}

SPE effectively enforces spatial alignment and transfers inter-frame consistency to the geometry estimation, leading to improved coherence between RGB conditions and predicted geometry maps.
Compared to channel concatenation, this method does not require modifying the input features of the denoising network, providing a more flexible fine-tuning mechanism. Ablation experiments~\ref{sec:ablation} demonstrate that this approach propagated the inherent inter-frame consistency more effectively, ensuring consistent and coherent geometric predictions across video frames.

\subsection{One-step Deterministic Training}
\label{sec:training_protocal}

Building on recent research on fine-tuning image diffusion models for geometric estimation~\cite{Garcia2024FineTuningID, xu2024diffusion}, we find that video diffusion models can also be fine-tuned as one-step deterministic models for geometric estimation. Following ~\cite{Garcia2024FineTuningID}, we no longer randomly sample timestep $t$ during training but instead fix $t=T$.
Additionally, we replace the Gaussian noise with its expectation, i.e., zero, and input it into the model along with the RGB latent representation. 
The video diffusion model is fine-tuned to match the latents of GT geometry attributes with an MSE loss. While significantly reducing computational cost, we find that the single-step model produces more accurate geometric predictions, as reported in Sec~\ref{sec:ablation}.

\section{Experiment}
\label{sec:experiment}

\begin{figure*}[t]
    \centering
    \includegraphics[width=0.95\linewidth]{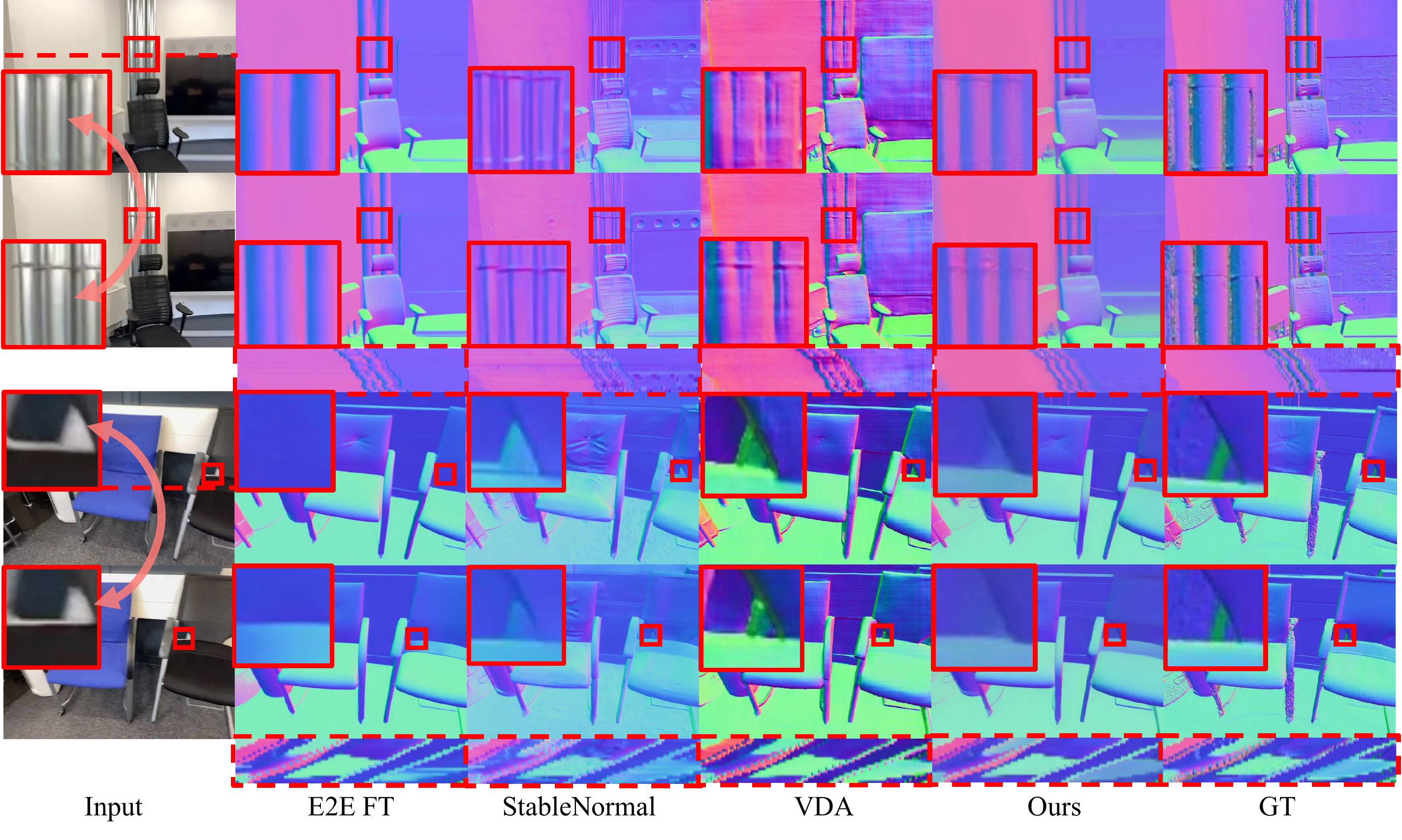}
    \vspace{-2mm}
    \caption{\textbf{Qualitative comparisons on normal estimation.} Our method produces more consistent normals, remains robust to highlight reflections, and generates results closest to the GT. It effectively removes noise from the GT, delivering smooth predictions.}
    \label{fig:video_normal_comparison_label}
    \vspace{-2mm}
\end{figure*}

\subsection{Setup}

\textbf{Dataset.} 
Following previous work~\cite{ye2024stablenormal, Ke2023RepurposingDI}, we trained our model exclusively on high-quality synthetic data. The datasets used for training include: Hypersim~\cite{Roberts2020HypersimAP}, an indoor multi-view dataset, using its position and normal labels, providing 40,000 samples after data grouping; InteriorNet~\cite{InteriorNet18}, an indoor video dataset, using its position and normal supervision, providing 30,000 samples after filtering; MatrixCity~\cite{li2023matrixcity}: an outdoor video dataset, using its normal data, providing 80,000 samples after filtering.

We choose the ScanNet++~\cite{yeshwanth2023scannet++} and 7scenes~\cite{Shotton2013SceneCR} dataset to evaluate the effectiveness of our method. Both are real-world scene datasets that has not been used during the training process. For Scannet++, the annotated geometric properties of each frame are re-rendered from a mesh scanned by a high-power LiDAR sensor (FARO Focus Premium laser scanner) and IMU camera poses, which can be used for depth and normal evaluation. 7scenes is used for the reconstruction point cloud evaluation.  We preprocess these two datasets into video clips for evaluation.

\textbf{Implementation Details.} Our model is fine-tuned based on CogVideoX~\cite{hong2022cogvideo} 5B, which employs RoPE~\cite{Su2021RoFormerET} for positional encoding. During training, we use the AdamW optimizer with a learning rate of 1e-4 and momentum parameters set to $\beta = (0.9, 0.999)$. The resolution of training video frames is (512, 384), and the batch size is set to 1. The entire training process runs for approximately 3 days on 8 A800 GPUs.


\subsection{Consistent Video Geometry Estimation}

Our method can estimate position and normal from a video clip. We conduct evaluations separately for these two aspects with current state-of-the-art methods. For image-based estiamtion methods, we choose Marigold~\cite{Ke2023RepurposingDI} and its normal version\footnote{https://huggingface.co/prs-eth/marigold-normals-lcm-v0-1}, E2E FT~\cite{Fu2024GeoWizardUT} and GeoWizard~\cite{fu2024geowizard}.  Additionally, we compare with the video-based geometric estimation method Video Depth Anything~\cite{video_depth_anything} (VDA). Since there are currently no existing works for estimating normals directly from videos, we recompute normals based on VDA’s depth results and camera intrinsic.
Note that all the aforementioned methods estimate geometric properties only in the local camera coordinate system. To enable a fair comparison with our approach, we first transform their results into the global coordinate system using the ground truth camera parameters before conducting the same evaluation.

\textbf{Normal Estimation.}
Following the metrics used in~\cite{ye2024stablenormal}, we compute the angular error between the predicted normals and the ground truth normals. We report both the mean and median angular errors, where lower values indicate higher accuracy. Additionally, we measure the percentage of pixels with an angular error less than 11.25°, where a higher value indicates better accuracy.

\begin{figure*}
    \centering
    \includegraphics[width=0.95\linewidth]{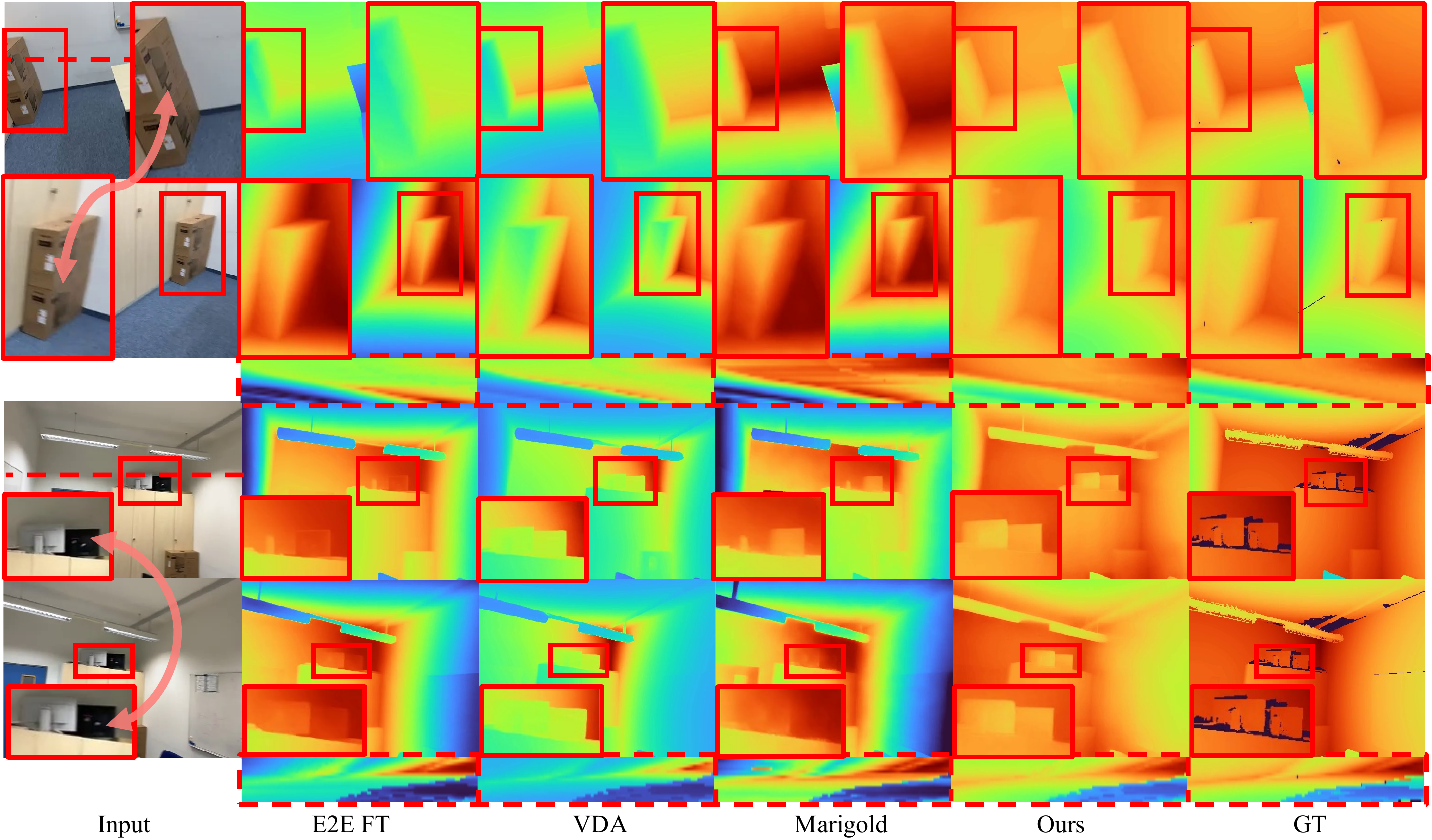}
    \vspace{-3mm}
    \caption{\textbf{Qualitative comparisons on radius estimation.} Our method achieves both accuracy and consistency, producing results closest to the ground truth. The entire video is normalized for consistent visualization.}
    \label{fig:videodepth_comparison_label}
    \vspace{-3mm}
\end{figure*}

\textbf{Radius Estimation.}
Note that depth is typically defined as the z-values of the 3D coordinate in the camera coordinate system. 
To convert it into consistent geometric properties, we use the distance from the 3D point to the origin of the global coordinate system (``radius'' in Fig~\ref{fig:pipeline}) as a substitute, which is aligned with GT by a least-square fitting. following~\cite{Ke2023RepurposingDI, video_depth_anything}. We report the  mean absolute relative error (AbsRel), defined as the average relative difference between the GT radius and the aligned counterpart at each pixel; the root mean square error (RMSE); and the percentage of pixels where the ratio of the
aligned predicted radius to the GT is less than 1.25 ($\delta_1$ accuracy).

\begin{table}[t]
  \centering
  \setlength\tabcolsep{5pt}
  \resizebox{0.95\linewidth}{!}{
    \begin{tabular}{@{\extracolsep{\fill}} l|SSS|SS }
        \toprule[1pt]
        \multirow{2}{*}{Methods} &
          \multicolumn{3}{c}{\textbf{Normal}} &
          \multicolumn{2}{c}{\textbf{Radius}} \\
          \cmidrule(lr){2-4}
          \cmidrule(lr){5-6}
          & {\textit{\small{Mean}}}$\downarrow$ & {\small{\textit{Median}}}$\downarrow$ & {\small{\textit{11.25}}}$\uparrow$ 
          & {\small{\textit{AbsRel}}}$\downarrow$ & 
          {\small{\textit{$\delta_1$}}}$\uparrow$\\
          \midrule
          Marigold~\cite{Ke2023RepurposingDI}
          & {20.93} & {11.36} & {53.31}
          & {11.2} & \cellcolor{second}{90.1} \\
          Geowizard~\cite{Fu2024GeoWizardUT}
          & {21.33} & {12.61} & {49.23}
          & {11.5} & {89.6} \\
            E2E FT~\cite{Garcia2024FineTuningID}
            & \cellcolor{second}{18.32} & \cellcolor{second}{8.22} & \cellcolor{best}{65.02} 
            & \cellcolor{best}{9.8} & {89.4} \\
          Stable Normal~\cite{ye2024stablenormal}
          & {23.51} & {13.16} & {61.68}
          & {/}  & {/} \\
          Video DA~\cite{video_depth_anything}
          & {28.54} & {18.70} & {49.78}
          & {13.5}  & {86.2} \\
          Ours
          & \cellcolor{best}{18.15} & \cellcolor{best}{7.91} & \cellcolor{second}{63.38}
          & \cellcolor{second}{10.2} & \cellcolor{best}{90.5} \\
        \bottomrule[1pt]
      \end{tabular}
      }
  \vspace{-2mm}
  \caption{\textbf{Evaluation on Scannet++ dataset.} Our method achieves state-of-the-art results in both normal and radius estimation.}
  \label{tab:video_normal_depth_comparition}
    \vspace{-5mm}
\end{table}

As shown in Table~\ref{tab:video_normal_depth_comparition}, our method achieves more consistent results in the global coordinate system compared to existing approaches, demonstrating its superior consistency in geometric estimation. Notably, unlike other methods, our approach does not require camera parameters as input, further proving the advantage of leveraging video diffusion model priors for this task.

\subsection{Video Reconstruction}
Since our method directly predicts the geometric properties of each frame's pixels in a unified coordinate system, it can be directly applied to reconstruction. We also compared our method with unposed image-based reconstruction methods, such as Dust3R~\cite{Wang2023DUSt3RG3} and Spann3R~\cite{Wang20243DRW}.

Dust3R estimates the coordinates of each image pair in their respective local frames, then aligns them through an optimization-based global alignment. Spann3R, on the other hand, maintains an external spatial memory and predicts subsequent frame coordinates based on existing status. In contrast, our method treats all frames as a single token sequence and directly predicts the global coordinates for all frames at once.
For a fair comparison, we remove Dust3R’s final alignment step in our evaluation. Please refer to the supplementary video materials for more comparisons.

\begin{figure}
    \centering
    \includegraphics[width=\linewidth]{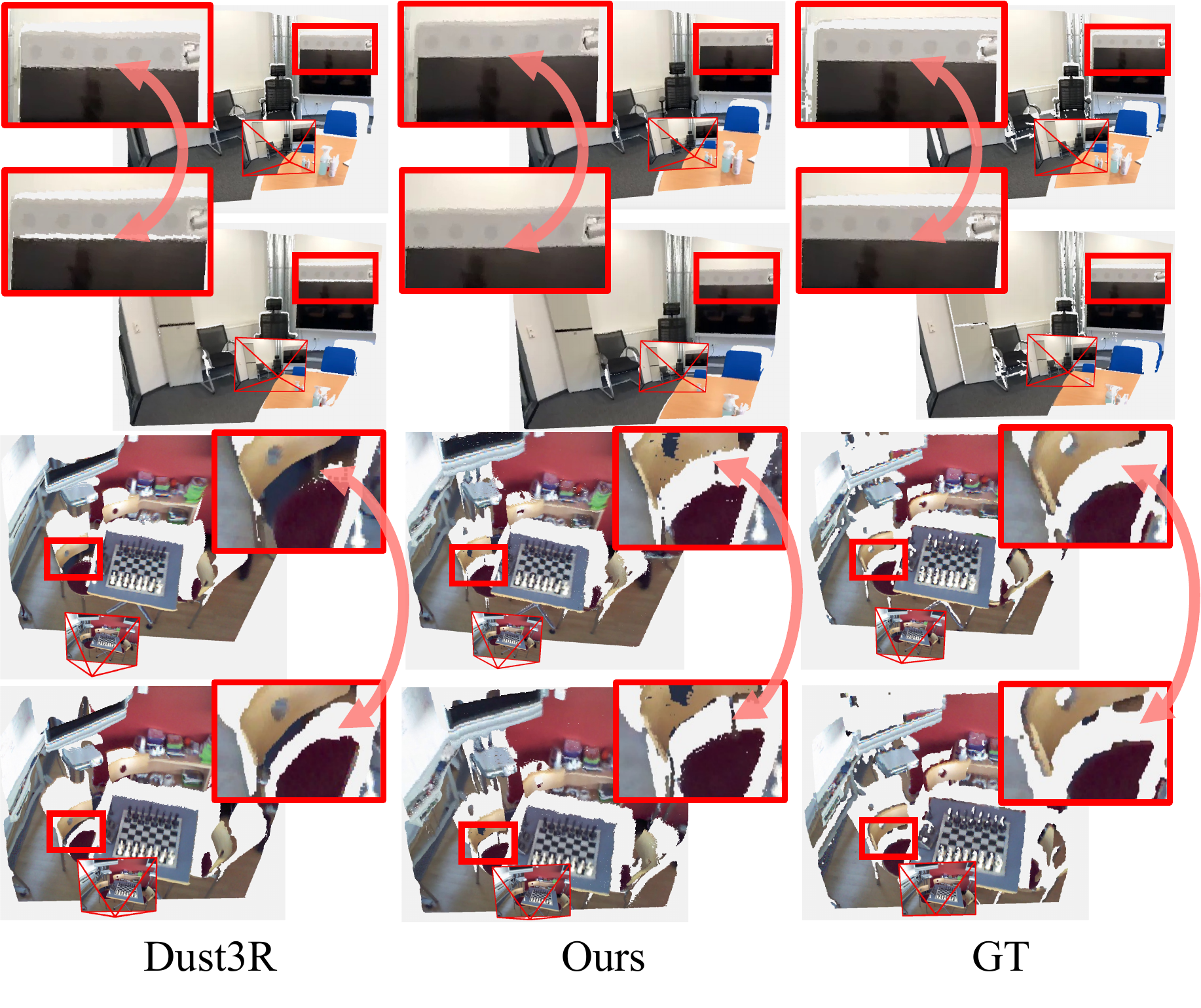}
    \vspace{-6mm}
    \caption{\textbf{Comparison to Dust3R.} Our method demonstrates better inter-frame consistency in multi-frame sequence reconstruction. Please refer to the videos in Supp for clearer comparisons. }
    \label{fig:video_recon_label}
    \vspace{-1mm}
\end{figure}

We reported the accuracy, completion and normal consistency, by directly comparing the predicted coordinate map with back-projected pixel depth. As shown in Table 2, our method achieves comparable performance to Spann3R and significantly outperforms Dust3R. Notably, our method is trained on far less data than the compared approaches, further demonstrating its potential.We also demonstrate in Fig~\ref{fig:video_recon_label} and the supplementary video materials that our method exhibits better inter-frame consistency compared to the baseline approaches.

\begin{table}[t]
  \centering
  \setlength\tabcolsep{5pt}
  \resizebox{0.95\linewidth}{!}{
    \begin{tabular}{@{\extracolsep{\fill}} l|SSS|SS}
        \toprule[1pt]
        \multirow{2}{*}{Methods} &
          \multicolumn{3}{c}{\textbf{Reconstruction}} &
          \multicolumn{2}{c}{\textbf{Depth}} \\
          \cmidrule(lr){2-4}
          \cmidrule(lr){5-6}
          & {\textit{\small{Acc}}}$\downarrow$ & {\small{\textit{Comp}}}$\downarrow$ & {\small{\textit{NC}}}$\uparrow$ 
          & {\small{\textit{AbsRel}}}$\downarrow$ & 
          {\small{\textit{$\delta1$}}}$\uparrow$\\
          \midrule
          Dust3R~\cite{Wang2023DUSt3RG3}
          & {0.216} & {0.073} & {0.573}
          & {25.8}  & {61.3} \\
          Spann3R~\cite{Wang20243DRW}
          & \cellcolor{best}{0.115} & \cellcolor{best}{0.038} & \cellcolor{best}{0.605}
          & \cellcolor{second}{21.70} & \cellcolor{second}{61.9} \\
            Ours
          & \cellcolor{second}{0.184} & \cellcolor{second}{0.058} & \cellcolor{second}{0.602}
          & \cellcolor{best}{20.69} & \cellcolor{best}{62.2} \\
        \bottomrule[1pt]
      \end{tabular}
      }
  \vspace{-2mm}
  \caption{\textbf{Evaluation on 7Scenes.} Our method achieves competitive results despite using significantly less training data.}\label{tab:video_recon}
    \vspace{-5mm}
\end{table}

\subsection{Dynamic Video Estimation}
Thanks to the rich external priors from video diffusion, our method can also be applied to consistent normal and depth estimation in dynamic scenes. We present some estimation results on DAVIS videos~\cite{pont20172017}, along with comparisons to single-frame estimation methods. For additional comparisons, please refer to the supplementary materials.

\subsection{Ablation Study}
\label{sec:ablation}
We evaluated the impact of different conditioning methods on the results in Table~\ref{tab:ablation}, including different condition methods : channel-wise concatenation (Chan Cat), sequential concatenation (Seq Cat), as well as our approach of reusing positional embeddings. 
We also compared the results of predicting a single attribute (w/o Multi-Attr) versus jointly predicting multiple attributes. 
Our findings show that optimizing multiple geometric properties together significantly improves performance.
Additionally, we compared with the results of predicting geometric properties in the seperate camera coordinate system of each frame (w/o Global Coord), This demonstrates that optimizing in a unified coordinate system better leverages inter-frame consistency, leading to improved performance. Please refer to the supplementary materials for visual comparison.


\begin{table}[t]
  \centering
  \setlength\tabcolsep{5pt}
  \resizebox{\linewidth}{!}{
    \begin{tabular}{@{\extracolsep{\fill}} l|SSSSS}
        \toprule[1pt]
        \multirow{2}{*}{Metric} & \multicolumn{5}{c}{\textbf{Variants}} \\
        \cmidrule(lr){2-6}
        & \textit{Chan Cat} & \textit{Seq Cat} & \textit{w/o Global Coord} & \textit{w/o Multi-Attr} & \textit{Ours} \\
        \midrule
        \textit{Mean}$\downarrow$ & \cellcolor{second}{19.13} & {19.75} & {19.91} & {20.29} & \cellcolor{best}{18.15} \\
        \textit{Median}$\downarrow$ & \cellcolor{second}{8.35} & {8.66} & {9.21} & {9.85} & \cellcolor{best}{7.91} \\
        \textit{11.25}$\uparrow$ & {61.17} & \cellcolor{second}{61.32} & {59.85} & {59.92} & \cellcolor{best}{63.38} \\
        \bottomrule[1pt]
      \end{tabular}
      }
  \vspace{-2mm}
  \caption{\textbf{Ablation study.} We ablate the key components in our design to show their effectiveness.}\label{tab:ablation}
  \vspace{-5mm}
\end{table}

\section{Limitation}
\label{sec:limitation}
Our method processes entire video clips but, due to storage constraints, can handle only limited-length segments per iteration. A significant challenge arises when stitching results from multiple short clips, as this can lead to accumulated errors. Integrating long-term memory into the current framework remains an open research problem. Additionally, the high computational cost restricts fine-tuning to lower resolutions (512$\times$384), occasionally causing blurry artifacts. Future work should explore efficient model distillation techniques to better capture inter-frame consistency, enhancing geometric predictions and overall output quality.

\section{Conclusion}
\label{sec:conclusion}
In this paper, we present \textit{UniGeo}, a unified framework to adapt pre-trained video generation models for consistent geometry estimation by leveraging their inherent inter-frame consistency. Specifically, we advocate optimizing geometric attributes in a global coordinate system rather than local camera coordinates, thus fully exploiting inter-frame priors encoded in pre-trained models. We further introduce a shared positional encoding method to precisely condition geometric attributes from RGB frames without modifying the network architecture. Additionally, our framework naturally integrates multiple geometric attributes in joint training, capitalizing on their shared correspondences to enhance overall performance. Extensive experiments demonstrate that our method effectively predicts consistent geometric attributes, with the resulting global geometry directly applicable to reconstruction tasks.

{
    \small
    \bibliographystyle{ieeenat_fullname}
    \bibliography{main}
}

\clearpage
\setcounter{page}{1}
\setcounter{section}{0}
\setcounter{figure}{0}
\setcounter{table}{0}
\maketitlesupplementary

\renewcommand{\thesection}{S\arabic{section}}
\renewcommand{\thesubsection}{S\arabic{subsection}}
\renewcommand{\thetable}{S\arabic{table}}
\renewcommand{\thefigure}{S\arabic{figure}}

\section{Outline}

In this supplementary file, we provide detailed description of multi-view data grouping, ablation comparison, and further results that could not be included in the main paper due to space constraints.

\section{Multi-view Data Processing}

Our primary dataset is Hypersim, where each scene consists of images captured from several discrete camera viewpoints. Therefore, we first preprocess the dataset by grouping the images. 


\begin{figure}[htb]
    \centering
    \includegraphics[width=0.95\linewidth]{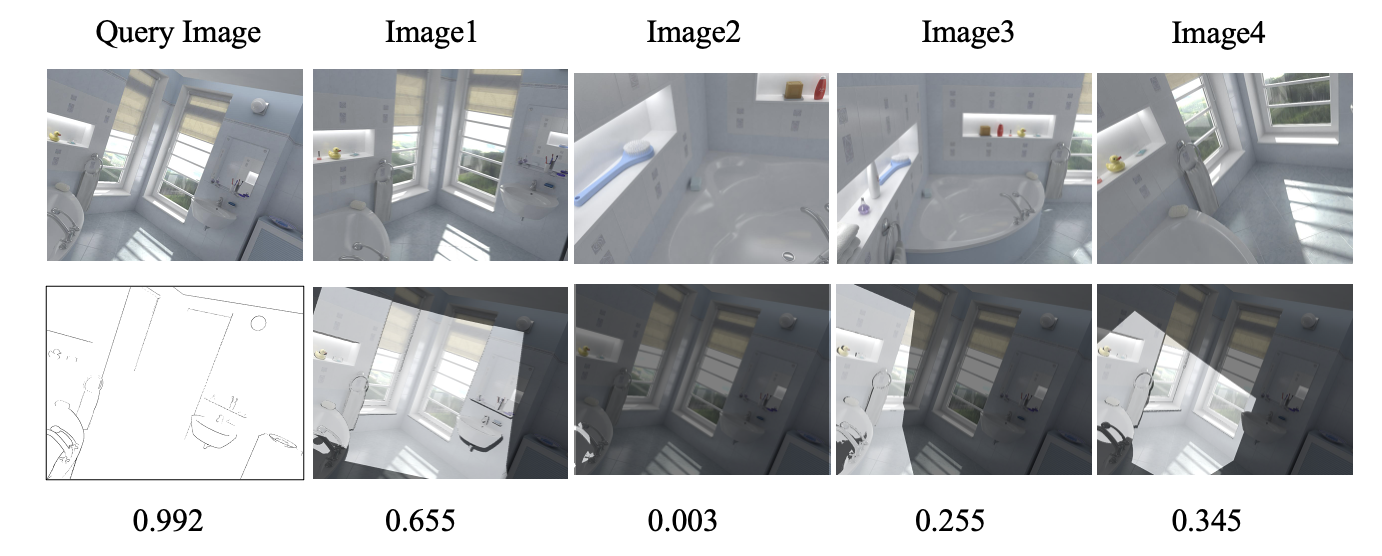}
    \caption{Given a query image, we use its depth and camera parameters to project it into other viewpoints. We then compute the overlap pixel ratio with the corresponding image to determine their relevance.}
    \label{fig:data_group}
\end{figure}

As illustrated by Fig~\ref{fig:data_group}, for each image in each scene, we we compute its relationship with any other images in this scene based on camera viewpoints and depth, denoted as $R_{ij}$. 

Specifically, for each scene, we compute the overlap between image pairs based on camera viewpoints and depth. For the $i-th$ image, we compute its bidirectional overlap with all other images using $(R_{ij} + R_{ji}) / 2$. We then select the top (NumView - 1) images with the highest overlap to form a data group with the i-th image for training.

\section{Coordinate Frame Definition}
UniGeo predicts geometric properties directly within a global coordinate space, thereby ensuring consistency for the same 3D point across different video frames. The global frame of reference is defined by the coordinate system of the first image in the video sequence.

To prepare the training data, given a sequence of $L$ frames with associated depth maps ${D_i}$, surface normals ${N_i}$, camera extrinsic parameters ${E_i}$, and intrinsic parameters ${K_i}$ ($i = 1, 2, \cdots, L$), we transform these local geometric properties into the global coordinate space as follows:

\begin{align}
\text{Coord}^{c}_i &= K_i^{-1}([U;V;\textbf{1}] \cdot D_i) \\
\text{Coord}^{g}_{i} &= E_1 E{i}^{-1} h(\text{Coord}^{{c}}_i) \\
\text{Normal}^{{g}}_i &= r(E_1 E{i}^{-1}) N_i
\end{align}

Here, $U$ and $V$ denote the pixel coordinate grids along the $x$ and $y$ axes, respectively; $h$ represents the transformation to homogeneous coordinates; and $r$ extracts the rotational component from the transformation matrix.

\section{Ablation Study Visualization}

In this section, we present visualizations of ablation experiments to demonstrate the effectiveness of our design choices.

\begin{figure*}[htb]
    \centering
    \includegraphics[width=0.95\linewidth]{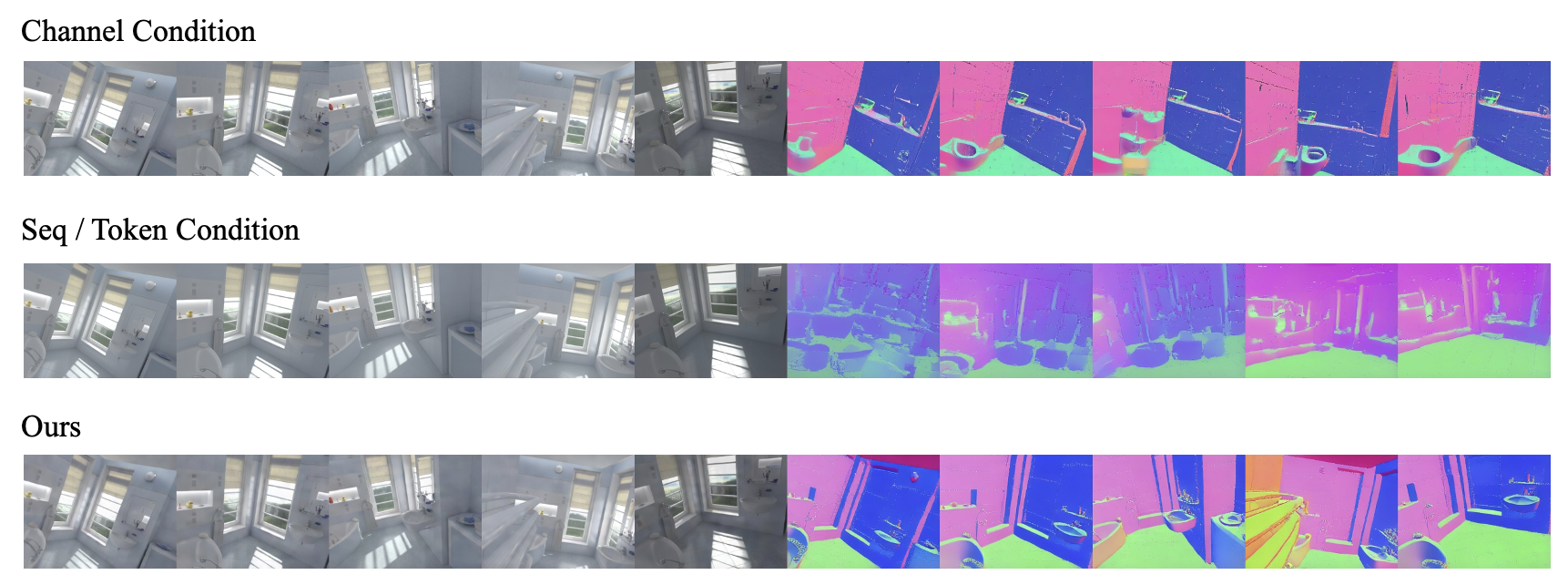}
    \caption{After the same number of fine-tuning steps, our method aligns RGB with geometric properties more quickly and produces more consistent predictions. This demonstrates that our proposed approach preserves the original prior as much as possible without disruption.}
    \label{fig:ablation_condation}
\end{figure*}

\textbf{Effectiveness of concatenation method.} We present a comparison of convergence speed across different concatenation methods. The evaluation is conducted on a small dataset consisting of 10 sequences, showing the results after 100 steps of training, as shown in Fig~\ref{fig:ablation_condation}. The results demonstrate that our proposed method, which reuses positional embeddings, achieves faster convergence, indicating that it effectively leverages the video diffusion prior. 

\begin{figure}[htb]
    \centering
    \includegraphics[width=\linewidth]{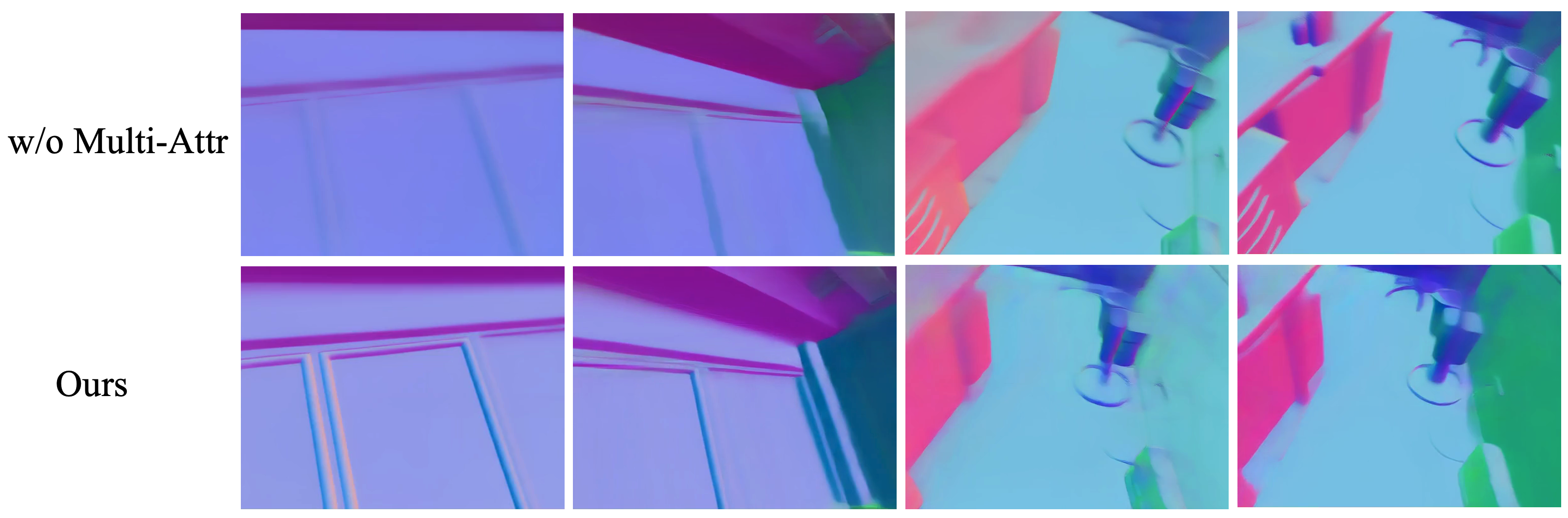}
    \caption{After the same number of fine-tuning steps, our method aligns RGB with geometric properties more quickly and produces more consistent predictions. This demonstrates that our proposed approach preserves the original prior as much as possible without disruption.}
    \label{fig:ablation_multi_attr}
\end{figure}

\textbf{Effectiveness of Multi-attributes Joint Training.}  We find that training multiple geometric properties simultaneously yields better results than training a single property in isolation. In Fig~\ref{fig:ablation_multi_attr}, we compare the normal estimation results when training only normal versus jointly training normal and position. The results show that joint optimization helps the model develop a better understanding of spatial geometry, leading to more reasonable predictions.

\textbf{Effectiveness of one-step deterministic training.}
In the realm of image-based geometry estimation tasks, works such as ~\cite{Garcia2024FineTuningID, xu2024matters} have achieved comparable or even superior performance by replacing the multi-step denoising generation of diffusion with a single-step process, while significantly reducing computational overhead. We have attempted to apply a similar approach to video diffusion models.

\begin{figure}[htb]
    \centering
    \includegraphics[width=0.95\linewidth]{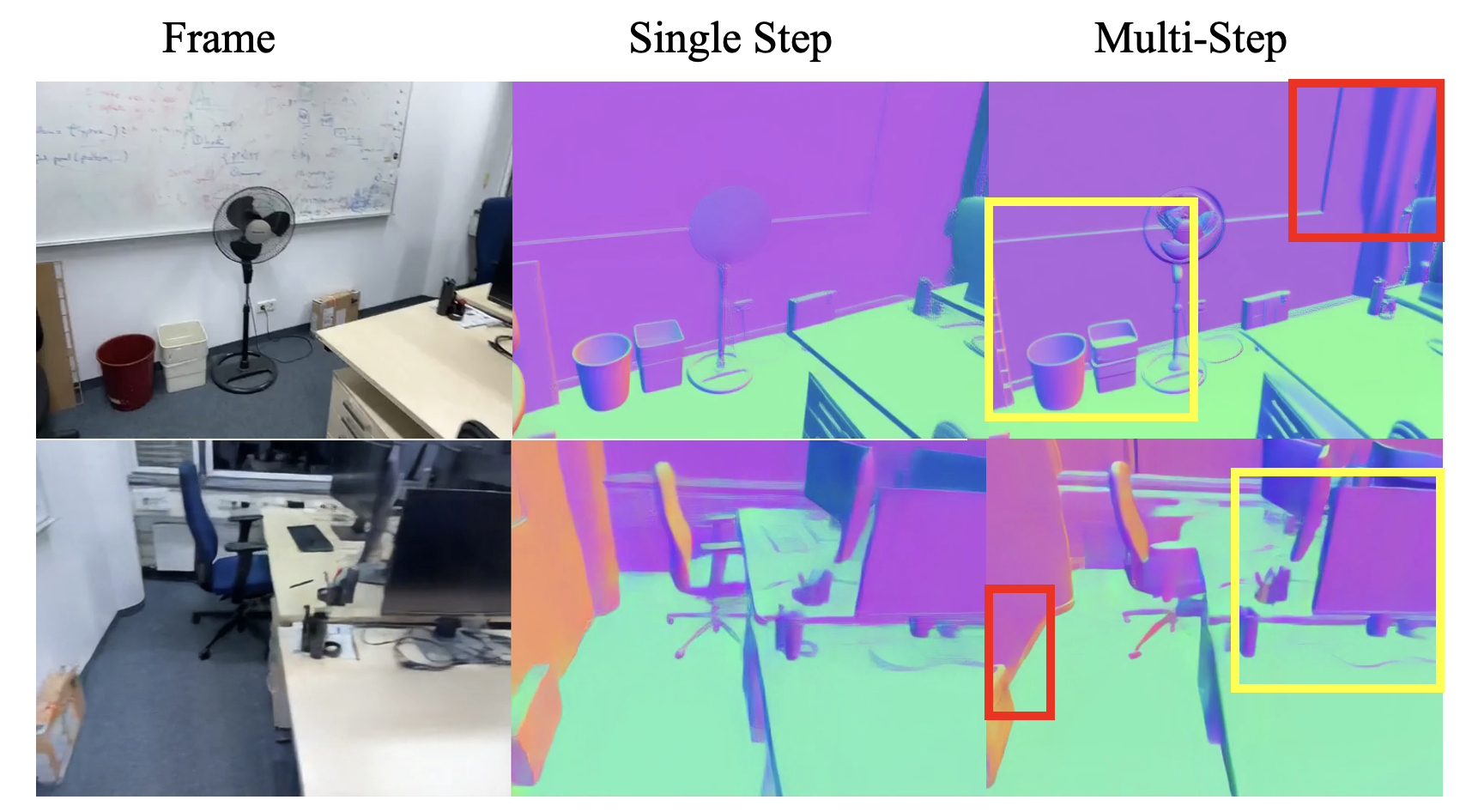}
    \caption{The figure illustrates a comparison between the results of multi-step and single-step approaches. The yellow boxes indicate areas where the multi-step method achieves sharper results, while the red boxes highlight regions where the multi-step approach exhibits prediction errors.}
    \label{fig:ablation_multistep}
\end{figure}

By fixing the timestep t at T during the training process and initializing the noise to the mean of a Gaussian distribution, i.e., zero, we have trained a single-step deterministic diffusion model for consistent geometric estimation. In Fig.~\ref{fig:ablation_multistep}, we present a comparison of the results between single-step and multi-step approaches under the condition of the same number of training steps.

Based on the results presented in Fig.~\ref{fig:ablation_multistep} and the numerical outcomes of the ablation study in the main text, it is evident that, overall, the one-step approach still holds a significant advantage in such deterministic prediction tasks.

\section{More Results}

In this section, we present additional experimental results, including  comparisons with current video depth estimation methods on additional datasets and more visualization.

\textbf{Comparison with video depth estimation methods.}
We further evaluate our approach on three static datasets (``ScanNet'', ``Neural RGBD'', and ``Replica'') and one dynamic dataset (``Bonn''), comparing it with SOTA video depth estimation methods. 
Here we report ``Depth" and ``Radius" metrics in ``\textbf{local}'' and ``\textbf{global}'' coordinate systems, respectively. Despite {being trained exclusively on limited static data}, our method still outperforms competitors and generalizes effectively to dynamic scenes, demonstrating the successful incorporation of the video diffusion model prior.

\begin{table}[ht]
  \vspace{-3.5mm}
  \centering
  \adjustbox{width={\linewidth},keepaspectratio}{
    \begin{tabular}{ll|cc|cc|cc|cc}
    \bottomrule
        \multirow{2}{*}{\textbf{Method}} & & \multicolumn{2}{c}{ScanNet} & \multicolumn{2}{c}{NRGBD} & \multicolumn{2}{c}{Replica} & \multicolumn{2}{|c}{Bonn} \\

         & & {\textit{\small{AbsRel}}}$\downarrow$ & {\small{{$\delta_1$}}}$\uparrow$ & 
            {\textit{\small{AbsRel}}}$\downarrow$ & {\small{{$\delta_1$}}}$\uparrow$ & 
            {\textit{\small{AbsRel}}}$\downarrow$ & {\small{{$\delta_1$}}}$\uparrow$ & 
            {\textit{\small{AbsRel}}}$\downarrow$ & {\small{{$\delta_1$}}}$\uparrow$  \\
        
        \hline
        \multirow{2}{*}{DepthCrafter}     & {local}   
        & \cellcolor{third}{7.28} & \cellcolor{third}{95.8} & {7.33} & {95.6} & \cellcolor{second}{6.21} & \cellcolor{second}{97.9} & \cellcolor{best}{7.47} & \cellcolor{best}{95.9} \\ 
        & {global} 
        & {7.35} & {95.3} & {7.52} & {94.3} & {6.41} & {97.3} & \cellcolor{second}{7.82} & \cellcolor{third}{95.2} \\ \hline
        \multirow{2}{*}{ChronoDepth}    & {local}   
        & {7.35} & {94.9} & \cellcolor{best}{6.02} & \cellcolor{best}{97.2} & {8.31} & {95.4} & {8.03} &  {94.6} \\
        & {global}   
        & {7.76} & {94.2} & \cellcolor{third}{6.91} & {96.4} & {8.65} & {94.9} & {8.24} &  {94.1} \\ \hline
        \multirow{2}{*}{DepthAnyVideo} & {local}   
        & {15.5} & {79.1} & {15.6} & {79.5} & {15.8} & {78.6} & {15.2} &  {78.3} \\
        & {global}   
        & {16.8} & {76.3} & {16.4} & {77.0} & {16.7} & {77.6} & {16.6} &  {76.9} \\ \hline
        \multirow{2}{*}{UniGeo} & {local}   
        & \cellcolor{second}{6.95} & \cellcolor{second}{95.9} & {7.14} & \cellcolor{third}{95.6} & \cellcolor{third}{6.35} & \cellcolor{third}{97.6} & {8.09} &  {95.1} \\
        & {global}   
        & \cellcolor{best}{6.69} & \cellcolor{best}{96.3} & \cellcolor{second}{6.58} & \cellcolor{second}{96.4} & \cellcolor{best}{5.53} & \cellcolor{best}{98.3} & \cellcolor{third}{8.03} &  \cellcolor{second}{95.6} \\
    \bottomrule
    \end{tabular}}
    \caption{Comparison with other video depth methods on additional datasets.}
    \label{table:depth}
  \vspace{-4.5mm}
\end{table}

\textbf{More visualization results.}
Here we visualize more comparison results. Please refer to the \href{https://sunyangtian.github.io/UniGeo-web/}{project page} for video results.

\begin{figure*}[htb]
    \centering
    \includegraphics[width=0.95\linewidth]{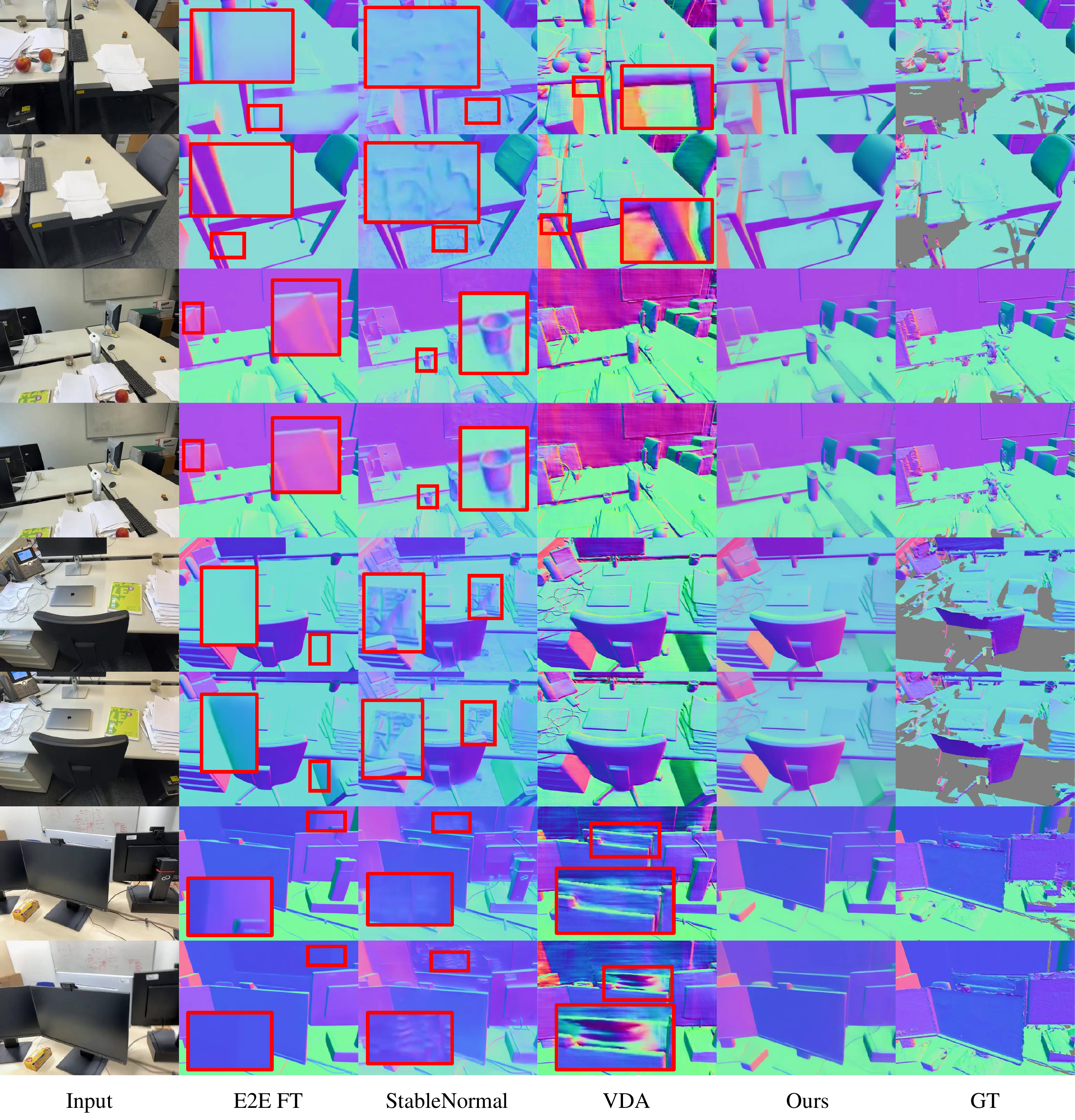}
    \caption{We show more visual comparisons of predicted normal on scannetpp dataset. The inconsistency is marked in red rectangles. It can be seen that ours achieve the most consistent visual effect while E2E and StableNormal provide inconsistent results, and VDA provides erroneous and inconsistent normals. }
    \label{fig:more_results1}
\end{figure*}

\begin{figure*}[htb]
    \centering
    \includegraphics[width=0.95\linewidth]{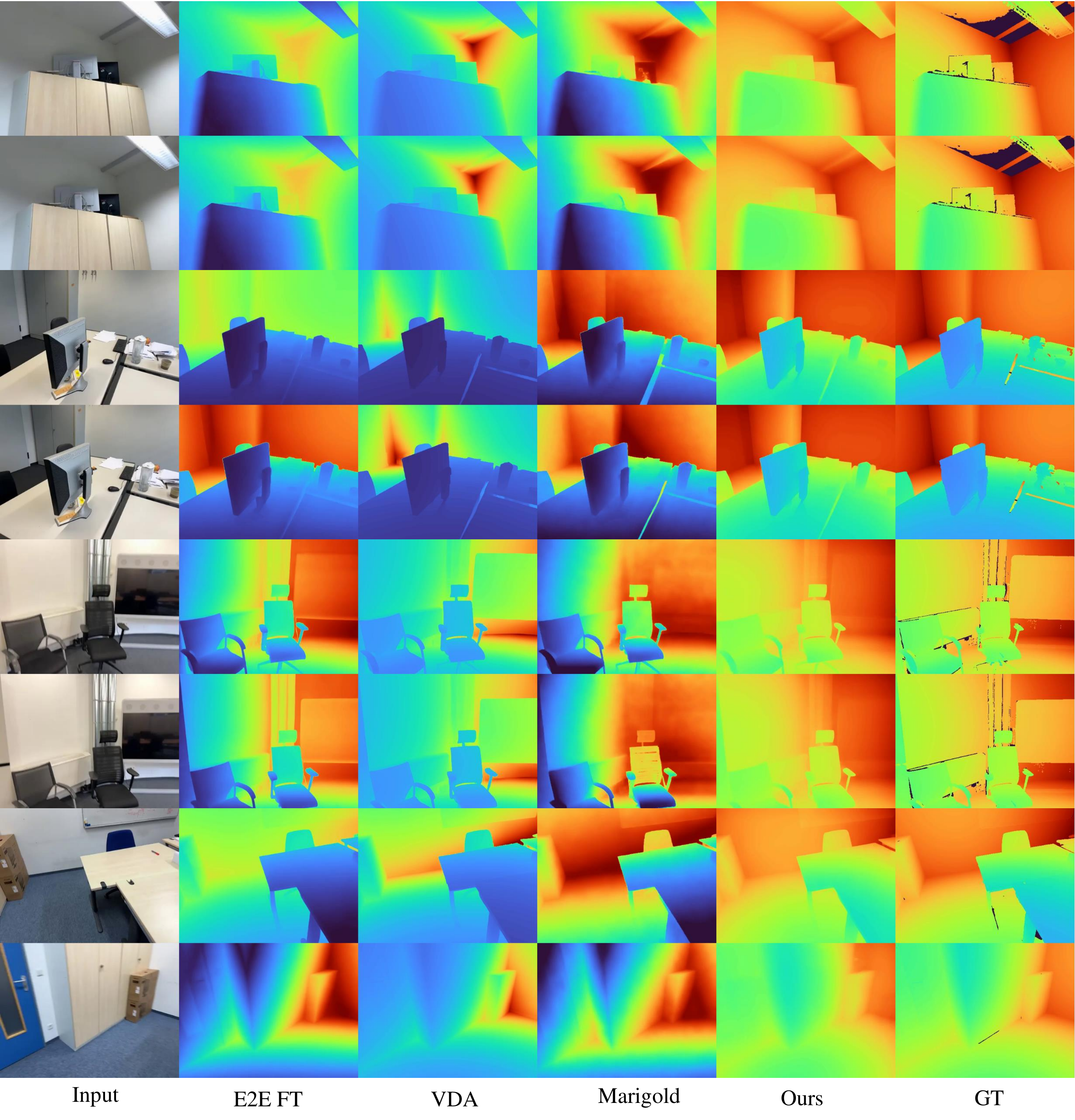}
    \caption{We show more visual comparisons of predicted radius on scannetpp dataset. Compared with other depth estimation methods, our approach produces more consistent and accurate geometry estimation.}
    \label{fig:more_results2}
\end{figure*}

\end{document}